\documentclass[letterpaper]{article} % DO NOT CHANGE THIS
\usepackage{aaai2026}  % DO NOT CHANGE THIS
\usepackage{times}  % DO NOT CHANGE THIS
\usepackage{helvet}  % DO NOT CHANGE THIS
\usepackage{courier}  % DO NOT CHANGE THIS
\usepackage[hyphens]{url}  % DO NOT CHANGE THIS
\usepackage{graphicx} % DO NOT CHANGE THIS
\usepackage{natbib}  % DO NOT CHANGE THIS AND DO NOT ADD ANY OPTIONS TO IT
\usepackage{caption} % DO NOT CHANGE THIS AND DO NOT ADD ANY OPTIONS TO IT
\usepackage{subcaption} % Added to support subfigure environment
\usepackage{algorithm}
\usepackage{algorithmic}
\usepackage{amsmath, amssymb, amsthm}
\usepackage{amsfonts}
\usepackage{pifont}
\usepackage{booktabs}
\usepackage{newfloat}
\usepackage{listings}
\DeclareCaptionStyle{ruled}{labelfont=normalfont,labelsep=colon,strut=off} % DO NOT CHANGE THIS
\floatstyle{ruled}
\newfloat{listing}{tb}{lst}{}
\floatname{listing}{Listing}
\newcommand{\algorithmicparfor}{\textbf{parallel for}}
\newcommand{\PARFOR}[1]{\STATE \algorithmicparfor\ #1\ \textbf{do} \begin{ALC@g}}
\newcommand{\ENDPARFOR}{\end{ALC@g}}

\makeatletter
\renewcommand{\PARFOR}[1]{\STATE \algorithmicparfor\ #1\ \textbf{do} \begingroup \let\COMMENT\ALC@com \begin{ALC@g}}
\renewcommand{\ENDPARFOR}{\end{ALC@g}\endgroup\STATE \textbf{end parallel for}}
\makeatother
\newcommand{\IP}{\textsc{IP}}
\newcommand{\TIR}{\textsc{TIR}}

\title{Q-ROAR: Outlier-Aware Rescaling for RoPE Position Interpolation in Quantized Long-Context LLMs}
\author{
     Ye Qiao, Sitao Huang
}
\affiliations{
\textit{University of California, Irvine,} Irvine, California, USA \\
    \{yeq6, sitaoh\}@uci.edu
}

\usepackage{bibentry}
\begin{document}

\maketitle
\begin{abstract}
Extending LLM context windows is crucial for long range tasks. RoPE-based position interpolation (PI) methods like linear and frequency-aware scaling extend input lengths without retraining, while post-training quantization (PTQ) enables practical deployment. We show that \emph{combining} PI with PTQ degrades accuracy due to coupled effects long context aliasing, dynamic range dilation, axis grid anisotropy, and outlier shifting that induce position-dependent logit noise. We provide the first systematic analysis of PI plus PTQ and introduce two diagnostics: \emph{Interpolation Pressure} (per-band phase scaling sensitivity) and \emph{Tail Inflation Ratios} (outlier shift from short to long contexts). To address this, we propose \textbf{Q-ROAR}, a RoPE-aware, weight-only stabilization that groups RoPE dimensions into a few frequency bands and performs a small search over per-band scales for $W_Q,W_K$, with an optional symmetric variant to preserve logit scale. The diagnostics guided search uses a tiny long-context dev set and requires no fine-tuning, kernel, or architecture changes. Empirically, Q-ROAR recovers up to 0.7\% accuracy on standard tasks and reduces GovReport perplexity by more than 10\%, while preserving short-context performance and compatibility with existing inference stacks.
\end{abstract}

\section{Motivation and Problem}
LLMs\cite{touvron2023llama} increasingly rely on long contexts for summarization, RAG, code, chain-of-thought, and argentic workflows. Inference-time RoPE scaling (PI)~\cite{peng2023yarn} extends the window is effective even without finetuning, but it is typically studied in full precision. Meanwhile, Post Training Quantization(PTQ)\cite{frantar2022gptq,lin2024awq} is essential for practical serving.

\begin{figure}[t]
\centering
\begin{subfigure}[t]{0.49\columnwidth}
    \centering
    \includegraphics[width=\linewidth]{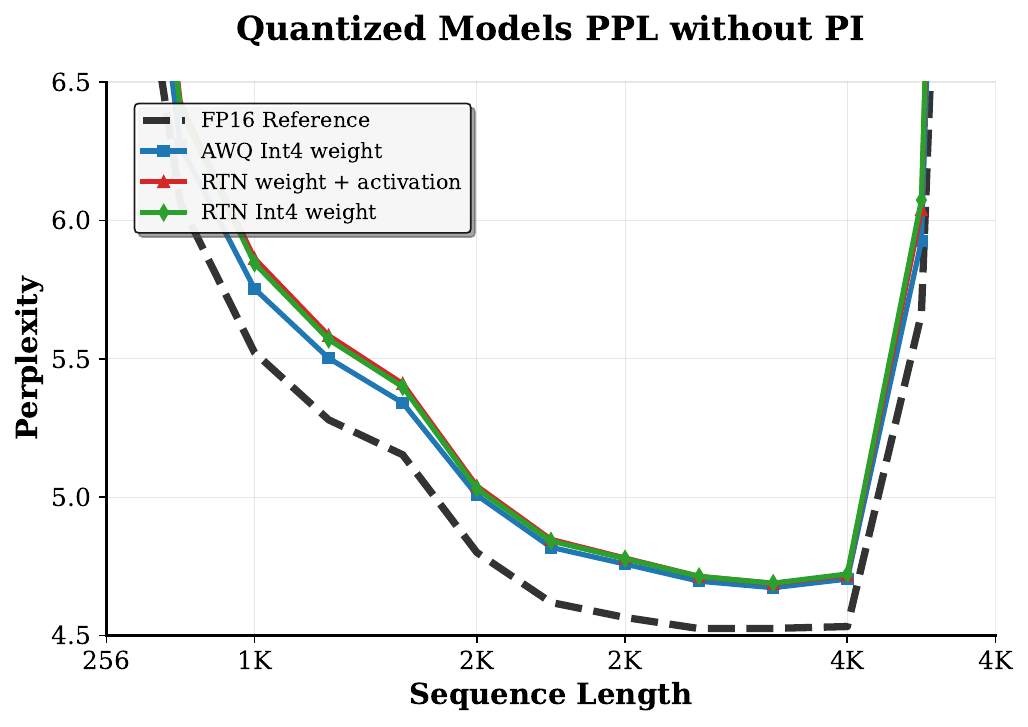}
    \caption{None Interpolated models}
    \label{left}
\end{subfigure}
\hfill
\begin{subfigure}[t]{0.5\columnwidth}
    \centering
    \includegraphics[width=\linewidth]{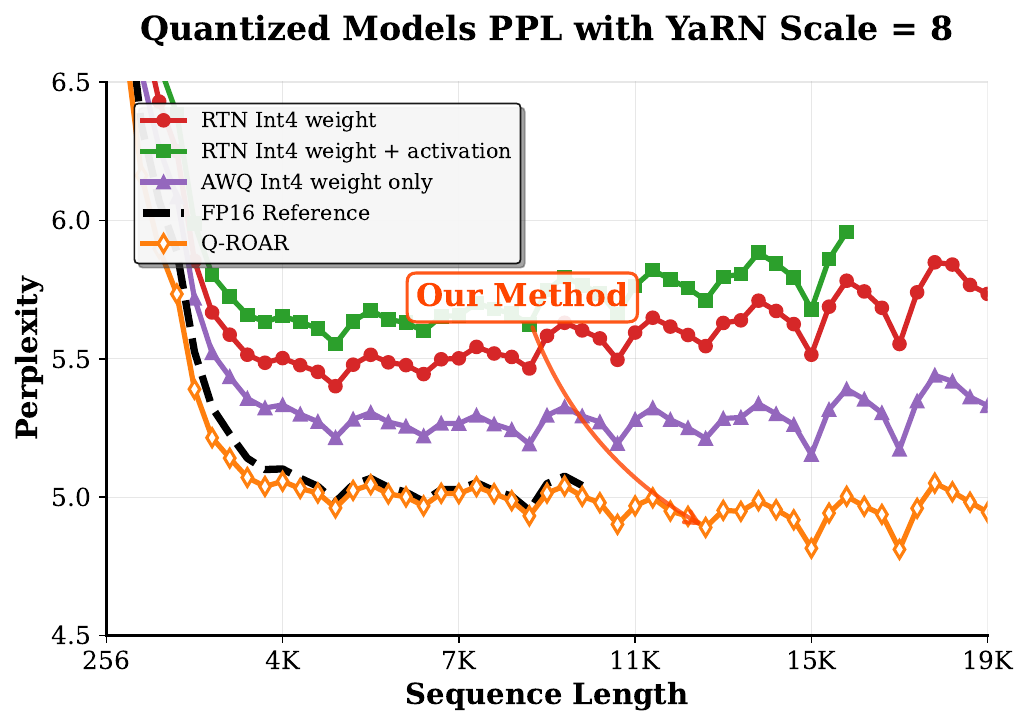}
    \caption{YaRN interpolation (s = 8)}
    \label{right}
\end{subfigure}
\caption{Quantized Llama-2-7b vs. GovReport PPL}
\label{fig:ppl}
\end{figure}

We observe that naively applying position interpolation (PI) to post-training quantized (PTQ) LLM models degrades accuracy both within and beyond the pretraining window (Figure~\ref{fig:ppl}). We compare quantized LLaMA-2-7B models with and without YaRN interpolation~\cite{peng2023yarn}. In the non-interpolated control (Figure~\ref{left}), quantized models behave as expected, showing only modest degradation. In contrast, under YaRN, all quantized variants deteriorate and most sharply for the generic RTN (round-to-nearest) configuration. AWQ performs better out of the box than RTN, suggesting that explicit activation outlier handling is implicated in the robustness gap. Motivated by this, we conduct a principled analysis and attribute the failures to \emph{coupling} between RoPE scaling and quantization: (i) \emph{aliasing} as high-frequency phases wrap; (ii) \emph{dynamic-range dilation} that inflates pre-activation tails; (iii) \emph{anisotropy} when axis-aligned quantizers operate on RoPE-rotated pairs; and (iv) \emph{outlier shift/amplification}. Together, these effects induce position-dependent logit noise.

\section{Interpolation Pressure and Tail Inflation}
Most RoPE scaling(interpolation) methods share a unified form:
\begin{equation}
\small
\phi_i^{\text{scaled}}(m)\;=\;\omega_i\,\frac{f(m)}{s_i},\quad s_i>0,
\label{eq:unified}
\end{equation}
where $f(\cdot)$ warps positions and $s_i$ rescales per-dimension frequency. Let the training regime support $|m-n|\!\le\!L_0$, target displacement $D$, and deviation
$\varepsilon_i(D)=\omega_i\!\left(\frac{f(D)}{s_i}-D_0\right)$. We define the sensitivity
\begin{equation}
\small
\IP_i \;=\; \left|\frac{\partial \varepsilon_i(D)}{\partial s_i}\right|=\omega_i\,\frac{f(D)}{s_i^2},
\end{equation}
which grows with $\omega_i$ and $D$, identifying high-frequency bands as fragile.

To capture PI-induced outlier shift, we use \emph{Tail-Inflation Ratios} at a high quantile $1-\varepsilon$:
\small
\begin{equation}
\tiny
\TIR^{\mathrm{W}}_i=\frac{Q_{|w_i^\top h|,\ \text{long}}(1-\varepsilon)}{Q_{|w_i^\top h|,\ \text{short}}(1-\varepsilon)}, \ \
\TIR^{\mathrm{A}}_i(m)=\frac{Q_{\|R(\phi_i^{\text{scaled}}(m))u_i\|_\infty}(1-\varepsilon)}{Q_{\|R(\phi_i(m))u_i\|_\infty}(1-\varepsilon)}.
\end{equation}

$\TIR^{\mathrm{W}}$ reflects pre-activation tail growth; $\TIR^{\mathrm{A}}$ reflects phase–axis amplitude inflation that increases activation clipping.

\paragraph{Geometric intuition.} RoPE preserves $\ell_2$ energy but rotates signal and quantization noise relative to axis-aligned bins; PI changes the angular trajectory over long spans. Thus, even uniform step sizes calibrated at short contexts become phase-misaligned at long contexts, increasing effective error along certain axes and perturbing attention logits.

\section{Q-ROAR (Band-wise Weight Rescaling)}
We partition RoPE pairs into $B$ log-spaced frequency bands $\{\mathcal{B}_b\}_{b=1}^B$ and assign a single scale $g_b$:
\begin{equation}
\small  
W_Q^{(b)} \!\leftarrow\! g_b\,W_Q^{(b)}, \ \
W_K^{(b)} \!\leftarrow\!
\begin{cases}
g_b\,W_K^{(b)} & \text{(shared mode)}\\
g_b^{-1}\,W_K^{(b)} & \text{(symmetric mode)}
\end{cases}
\end{equation}

% Symmetric mode approximately preserves logit magnitude. We derive per-band windows $\mathcal{G}_b=[g_b^{\min},g_b^{\max}]$ by combining \IP\ (tight bounds for high frequencies) with $\TIR^{\mathrm{W}}$ (mild shrinkage for inflated tails). A tiny long-context dev set (e.g., $\sim$10 docs) drives a lightweight grid search that minimizes a length-weighted perplexity objective.

% \paragraph{Parameter Choices}
% \begin{itemize}
% \item \textbf{Band count:} $B\in\{6,8\}$; map pairs by log-frequency to keep bands compact and stable.
% \item \textbf{Search grid:} 5–9 log-spaced candidates per band; coordinate or joint grid search; early-stop when validation improves $<\eta$.
% \item \textbf{Objective:} $\sum_{L\in\mathcal{L}} w_L\,\mathrm{ppl}(L;\{g_b\})$ with $w_L$ emphasizing longer lengths; reuse KV where possible.
% \item \textbf{Safety:} Clip $g_b$ by $\gamma_b=1+\frac{\tau}{1+\log(\omega_{b,\mathrm{med}}/\omega_{\min})}$ and $\kappa/\TIR^{\mathrm{W}}_b$ (\(\kappa\!\in[1.0,1.3],\,\tau\!\in[0.2,0.5]\)).
% \end{itemize}
Symmetric mode approximately preserves logit magnitude. We set per-band windows $\mathcal{G}_b=[g_b^{\min},g_b^{\max}]$ by combining \IP\ (tight for high frequencies) with $\TIR^{\mathrm{W}}$ (shrink inflated tails). A tiny long-context dev set ($\sim$10 docs) drives a lightweight grid search that minimizes a length-weighted perplexity objective.

We choose \textbf{Band count} $B\!\in\!\{6,8\}$ with log-frequency grouping; 
\textbf{Grid} 5–9 log-spaced candidates per band with coordinate or small joint search and early stopping (gain $<\eta$); 
\textbf{Objective} $\sum_{L\in\mathcal{L}} w_L\,\mathrm{ppl}(L;\{g_b\})$ with $w_L$ emphasizing longer lengths and KV reuse; 
\textbf{Safety} bound $g_b$ via $\gamma_b=1+\frac{\tau}{1+\log(\omega_{b,\mathrm{med}}/\omega_{\min})}$ and $\kappa/\TIR^{\mathrm{W}}_b$, with $\kappa\!\in\![1.0,1.3]$, $\tau\!\in\![0.2,0.5]$.

We focus on rescaling \emph{weights} of key and query projection layers ($W_Q,W_K$) rather than adjusting activation quantization for three reasons. (i) Activation statistics drift with content/position under PI; changing activation clips/steps couples to kernels/runtimes. (ii) Weight rescaling is static, quantizer and kernel agnostic (AWQ/RTN), and portable. (iii) Symmetric scaling keeps logit scale stable, avoiding retuning.

\begin{algorithm}[t]
\small
\caption{Q-ROAR search}
\label{alg:qroar}
\begin{algorithmic}[1]
\STATE Partition RoPE dims into $\{\mathcal{B}_b\}_{b=1}^B$
\STATE Estimate per-band $\IP_b$ and $\TIR^{\mathrm{W}}_b$ from short vs.\ PI long-context caches
\STATE Derive band windows $\mathcal{G}_b$; build log-spaced grids $G_b\subset\mathcal{G}_b$
\STATE \textbf{for} candidates $\{g_b\in G_b\}$ \textbf{do} evaluate ppl on tiny dev set
\STATE Select $\{g_b^\star\}$ minimizing the objective; prefer symmetric mode; fallback to shared if unstable
\STATE Serialize $\{g_b^\star\}$ and mode in model metadata
\end{algorithmic}
\end{algorithm}

% \section{Related Work and Positioning}
% RoPE scaling methods (Linear~\cite{chen2023extending}, YaRN~\cite{peng2023yarn}, LongRoPE~\cite{ding2024longrope}) target full-precision extrapolation; PTQ methods (GPTQ~\cite{frantar2022gptq}, SmoothQuant~\cite{xiao2023smoothquant}, AWQ~\cite{lin2024awq}) reduce memory/latency but are calibrated on short contexts. Our contribution is orthogonal: we analyze the \emph{interaction} between PI and PTQ, introduce diagnostics (\IP, \TIR) that expose coupling, and convert them into a weight-only, band-limited rescaling that preserves compatibility with existing inference stacks.

We sampled 10 long documents from Proof-pile \cite{Proofpile} dataset that each of them are longer than 60k tokens; cache short vs.\ PI distributions to compute \TIR. \textbf{Complexity.} with running average sliding window size of 256. If each band has $K$ candidates, evaluation cost is $\mathcal{O}(BK)$ runs (coordinate search) or $\mathcal{O}(K^B)$ (small $B$ only) with token reuse via sliding windows. \textbf{Serialization.} Store $\{g_b\}$ alongside checkpoint; apply at load time to $(W_Q,W_K)$ tensors only. The search was conducted using two NVIDIA RTX 4090 and roughly took 4 GPU hours.

\section{Experiments}\label{sec:exp}
For long-context evaluation, we benchmarked GovReport, and for standard LLM performance we used WikiText2\cite{wikitext} and five zero-shot tasks. At the base 4K context, quantized models track FP16 with only modest degradation. Under YaRN interpolation (32K/64K), RTN and AWQ exhibit clear drops in accuracy and higher perplexity, confirming the destructive interaction between position interpolation and quantization. In contrast, Q-ROAR consistently improves robustness that recovering up to 0.7\% accuracy on the standard suite (notably, even where PI–PTQ coupling is minimal) and yielding over 10\% relative perplexity reductions on GovReport versus RTN while preserving short-context performance. Overall, Q-ROAR mitigates aliasing and outlier amplification, enabling stable long context inference in quantized LLMs without retraining or kernel changes.

% \textbf{\TODO{Insert setup, datasets (e.g., GovReport/PG19/LongBench), PI schemes and lengths, PTQ baselines (GPTQ/AWQ/RTN), metrics, and main table/figure.}}\\
% \noindent\textbf{Key Results (to be populated).} \TODO{Summarize $>\!10\%$ gains, negligible overhead, and ablations on $B$, symmetric vs.\ shared, and diagnostics ablations.} 

\begin{table}[t]
\centering
\caption{GovReport perplexity on LLaMA-2-7B across evaluation context sizes (lower is better).}
\label{tab:govreport-multisize}
\resizebox{0.7\linewidth}{!}{%
\begin{tabular}{lcccccc}
\toprule
\textbf{Setting} & \textbf{Context} & \multicolumn{5}{c}{\textbf{Evaluation Context Window Size}} \\
 &  & 2048 & 4096 & 8192 & 16384 & 32768 \\
\midrule
\multicolumn{7}{c}{\emph{Extended Context with YaRN (64K tokens, $s{=}16$)}} \\
\midrule
FP16       & 64K & 4.437 & 4.359 & 4.329 & 4.175 & 6.069 \\
RTN W4     & 64K & 4.544 & 4.485 & 4.470 & 4.485 & 6.713 \\
AWQ W4     & 64K & 4.489 & 4.421 & 4.405 & 4.414 & 6.302 \\
\textbf{Q-ROAR W4} & \textbf{64K} & \textbf{4.444} & \textbf{4.393} & \textbf{4.321} & \textbf{4.181} & \textbf{5.833} \\
\bottomrule
\end{tabular}%
}
\end{table}

\begin{table}[t]
    \centering
    \caption{Performance of LLaMA-2-7B on standard LLM benchmarks under different quantization and PI settings.}
    \label{tab:result-llama-2-7b}
    \resizebox{0.35\textwidth}{!}{
    \begin{tabular}{lccc}
        \toprule
        \textbf{Setting} & \textbf{Context} & \textbf{5-shot Avg.} & \textbf{WikiText2 PPL} \\
        \midrule
        \multicolumn{4}{c}{\emph{Base Context (4096 context window)}} \\
        \midrule
        Baseline FP16          & 4k & 70.83\% & 5.47 \\
        RTN W4                 & 4k & 64.14\% & 5.73 \\
        RTN W4-A4              & 4k & 62.02\% & 5.92 \\
        AWQ W4                 & 4k & 64.25\% & 5.61 \\
        AWQ W4-A4              & 4k & 62.31\% & 5.77 \\
        \midrule
        \multicolumn{4}{c}{\emph{Extended Context with YaRN (32K context window, $s{=}8$)}} \\
        \midrule
        Baseline FP16          & 32K & 63.71\% & 6.09 \\
        RTN W4                 & 32K & 63.32\% & 6.41 \\
        RTN W4-A4              & 32K & 62.84\% & 6.60 \\
        AWQ W4                 & 32K & 63.52\% & 6.31 \\
        AWQ W4-A4              & 32K & 62.53\% & 6.49 \\
        \midrule
        \textbf{Q-ROAR W4}     & 32K & \textbf{63.96\%} & \textbf{6.19} \\
        \textbf{Q-ROAR W4-A4}  & 32K & \textbf{63.21\%} & \textbf{6.40} \\
        \bottomrule
    \end{tabular}
    }
\end{table}

\section{Conclusion}
We analyzed why PI harms PTQ LLMs and introduced \IP\ and \TIR\ to quantify the coupling between RoPE scaling and quantization. Building on these diagnostics, \textbf{Q-ROAR} provides a portable, weight-only, bandwise rescaling of $(W_Q,W_K)$ that delivers consistent long context gains with negligible overhead and no kernel changes. While Q-ROAR assumes RoPE and targets weight-quantized models (heavy activation/KV quantization may still benefit from modest clip expansion guided by \TIR), it substantially mitigates aliasing and outlier amplification at extended lengths. Future work includes extending the approach to non-RoPE position encoding and integrating lightweight, context-aware activation calibration.

\bibliography{aaai2026}

\begin{thebibliography}{6}
\providecommand{\natexlab}[1]{#1}

\bibitem[{Frantar et~al.(2022)Frantar, Ashkboos, Hoefler, and Alistarh}]{frantar2022gptq}
Frantar, E.; Ashkboos, S.; Hoefler, T.; and Alistarh, D. 2022.
\newblock Gptq: Accurate post-training quantization for generative pre-trained transformers.
\newblock \emph{arXiv preprint arXiv:2210.17323}.

\bibitem[{Lin et~al.(2024)}]{lin2024awq}
Lin, J.; et~al. 2024.
\newblock Awq: Activation-aware weight quantization for on-device llm compression and acceleration.
\newblock \emph{Proceedings of machine learning and systems}, 6: 87--100.

\bibitem[{Merity et~al.(2016)}]{wikitext}
Merity, S.; et~al. 2016.
\newblock Pointer sentinel mixture models.
\newblock \emph{arXiv preprint arXiv:1609.07843}.

\bibitem[{Peng et~al.(2023)Peng, Quesnelle, Fan, and Shippole}]{peng2023yarn}
Peng, B.; Quesnelle, J.; Fan, H.; and Shippole, E. 2023.
\newblock Yarn: Efficient context window extension of large language models.
\newblock \emph{arXiv preprint arXiv:2309.00071}.

\bibitem[{Touvron(2023)}]{touvron2023llama}
Touvron, H.~o. 2023.
\newblock Llama 2: Open foundation and fine-tuned chat models.
\newblock \emph{arXiv preprint arXiv:2307.09288}.

\bibitem[{Zhangir~Azerbayev(2022)}]{Proofpile}
Zhangir~Azerbayev, B.~P., Edward~Ayers. 2022.
\newblock Proof-pile.

\end{thebibliography}

\end{document}